\documentclass[12pt]{article}

\usepackage{fontspec}
\usepackage[
  letterpaper,
  inner=1.25in,
  outer=1in,
  top=1in,
  bottom=1.25in,
  marginparwidth=0.6in,
]{geometry}

\usepackage[british]{babel}
\usepackage[final]{microtype}

\usepackage{titlesec}
\titleformat{\section}
  {\normalfont\scshape}
  {\llap{\thesection\quad}}
  {0pt}
  {}
\titleformat{\subsection}
  {\normalfont\normalsize\scshape}
  {\thesubsection\quad}
  {0pt}
  {}
\titleformat{\subsubsection}
  {\normalfont}
  {\thesubsubsection\quad}
  {0pt}
  {}
\titlespacing*{\section}{0pt}{2ex plus 1ex minus .2ex}{1ex plus .2ex}
\titlespacing*{\subsection}{0pt}{1.5ex plus 1ex minus .2ex}{0.5ex plus .2ex}
\titlespacing*{\subsubsection}{0pt}{1ex plus 0.5ex minus .1ex}{0.3ex plus .1ex}

\usepackage{fancyhdr}
\renewcommand{\headrulewidth}{0pt}
\usepackage{xcolor}
\definecolor{linkmaroon}{RGB}{128, 0, 32}
\usepackage{hyperref}
\hypersetup{
  colorlinks=true,
  linkcolor=linkmaroon,
  citecolor=linkmaroon,
  urlcolor=linkmaroon,
  pdfauthor={Brett Reynolds},
}
\usepackage{orcidlink}

\usepackage[style=american]{csquotes}
\MakeOuterQuote{"}

\newcommand{\term}[1]{\textsc{#1}}
\newcommand{\mention}[1]{\textit{#1}}

\usepackage{marvosym}

\usepackage{amsmath,amssymb}
\usepackage{langsci-gb4e}
\makeatletter
\@ifundefined{noautomath}{}{\noautomath}
\makeatother

\usepackage{enumitem}
\setlist{nosep, leftmargin=*}
\setlist[enumerate]{label=\arabic*.}
\setlist[itemize]{label=--}

\usepackage{booktabs}
\usepackage{array}
\usepackage{graphicx}
\graphicspath{{figures/}}

\usepackage[backend=biber,style=apa,natbib=true,doi=true,isbn=false,url=true]{biblatex}
\newcommand{\posscite}[1]{\citeauthor{#1}'s (\citeyear{#1})}

\usepackage{xspace}

\usepackage{float}
\usepackage{tikz}
\usetikzlibrary{arrows.meta}
\tikzset{
  ttpbox/.style={draw, rounded corners=2pt, align=center, inner sep=4pt,
                 minimum height=7.5mm, font=\small},
  ttpfigbox/.style={ttpbox, inner sep=3pt, minimum height=6mm, font=\footnotesize},
  ttpar/.style={-{Stealth[length=2mm]}, thick},
  ttpblocked/.style={dashed, thick},
}

\hypersetup{
  pdftitle={Grounding Without Corrective Control: Truth-Tracking Profiles for Large Language Models},
  pdfkeywords={large language models, grounding, answerability, corrective control, projectibility, truth-tracking}
}

\newif\ifttpanonymous
\ifdefined\TTPANONYMOUS
  \ttpanonymoustrue
  \hypersetup{pdfauthor={}}
\else
  \ttpanonymousfalse
\fi

\providecommand{\aidisclosure}[1]{%
  \par\medskip
  \noindent{\small\textsc{AI use.}\enspace The large language models #1 served as
  drafting and editing aids throughout the preparation of this paper. I'm
  responsible for all theoretical claims, arguments, errors, and interpretive
  choices.\par}%
  \medskip
}

\newcommand{\transitionpagehead}[1]{%
  \fancypagestyle{ttptransition}{%
    \fancyhf{}%
    \fancyhead[L]{\small\scshape #1}%
    \fancyhead[R]{\small\thepage}%
    \renewcommand{\headrulewidth}{0pt}%
  }%
  \thispagestyle{ttptransition}%
}

\title{Grounding Without Corrective Control:\\ Truth-Tracking Profiles for Large Language Models}
\ifttpanonymous
\author{}
\date{}
\else
\author{Brett Reynolds \orcidlink{0000-0003-0073-7195}%
\thanks{Contact: \href{mailto:brett.reynolds@humber.ca}{brett.reynolds@humber.ca}}\\
Humber Polytechnic \& University of Toronto}
\date{14th August 2026}
\fi

\begin{document}
\maketitle

\begin{abstract}
Recent work suggests that some large language model representations
have content or reference. Grounding can secure either without supplying live routes for
correction. This paper asks what follows from that gap. An output is \term{answerable} when
discrepancies can affect what a target- and task-specific \term{arrangement} produces,
accepts, or withdraws. The arrangement has \term{corrective control} only when live,
sufficiently independent routes can detect and repair fresh discrepancies. A route
\term{profile} records which routes constrain the arrangement and how they're related.
Those profiles support analysis of \term{truth-tracking}: patterned support for
representational success.

Language models are the pressure case; text-only arrangements provide a task-relative
limiting case. Text-trained models inherit patterns of testimony, coherence, and
prior correction. Where target-sensitive correction survives training, these can supply
\term{derivative answerability} (inherited constraint); \term{live answerability} is the
relation supplied by a current route for fresh discrepancies. Fluent failures should
follow when a task requires independently informative access to the facts. Self-consistency,
retrieval, tools, code execution, multimodal input, and feedback should help selectively.
Route-by-task interactions test the distinctions. The decomposition's empirical burden is
to predict held-out route--task combinations or improve intervention choice without
conceptual refitting. Surface improvement and truth-tracking improvement can come apart.
\end{abstract}

\noindent\textbf{Keywords:} large language models; grounding; answerability; corrective
control; projectibility; truth-tracking

\ifttpanonymous
  \aidisclosure{Claude and Codex}
\else
  \aidisclosure{Anthropic Claude Opus 4.8, Fable 5, and Opus 5; and OpenAI Codex using
  GPT-5.5}
\fi

\section{Introduction: from grounding to corrective control}\label{sec:intro}

Do the representations inside large language models (LLMs) connect to the world? The
question is often put as a yes/no. \textcite{bender2020} say no. They imagine two
stranded people communicating by underwater cable. An octopus that can tap the cable
but can't observe either island learns to predict their messages and attempts to impersonate
one correspondent. Because it has access only to linguistic form, they argue, its responses
remain disconnected from what the speakers discuss. Others reply that such systems
meet some conditions on reference, or that sensory grounding can be optional
for thought in the first place
\citep{coelho_mollo_milliere_2026_vector_grounding_problem,chalmers_2023_thought_sensory_grounding_llms}.

But that opposition only starts the inquiry. The debate descends from the symbol-grounding
problem
\citep{harnad_1990_symbol_grounding_problem}: how can symbols, or symbol-like states,
get meaning independent of the interpretations supplied by users? For LLMs, recent work
has pluralized the issue. \textcite{pavlick_2023_symbols_and_grounding_llms}
treats grounding as a family of questions about symbols and use;
\textcite{coelho_mollo_milliere_2026_vector_grounding_problem} distinguish kinds of
grounding; and \textcite{chalmers_2023_thought_sensory_grounding_llms} separates sensory
grounding from thought.

The recent literature has answered sceptical arguments one at a time, principally on the
side of content. On reference, the natural histories of a language model's words can
secure it \citep{mandelkern_linzen_2024_do_language_models_words_refer}, while intentions
are the wrong place to look \citep{pepp_2025_reference_without_intentions_llms}. On
linguistic competence and meaning, the argument from communicative intention fails at
both premises \citep{attah_2025_do_language_models_lack_communicative_intentions}, and
facts about linguistic practice can determine meaning even if the facts that would
determine mental content don't
\citep{grindrod_2024_large_language_models_linguistic_intentionality}. At the level of the
proposed target, meaning can be grounded in a corpus rather than in the world
\citep{havlik_2024_meaning_understanding_llms_synthese}.
\textcite{ruyant_2026_what_do_llms_represent} reaches the further conclusion that what such
systems represent is a class of appropriate linguistic production rather than the world.

I contest none of those claims, and my argument doesn't require their authors to be
mistaken. Some of those authors discuss misrepresentation, error, and repair, but the
grounding accounts considered here don't by themselves supply an account of the live,
task-indexed routes through which a fresh discrepancy can be detected and translated into
a target-improving revision.\footnote{%
\textcite{coelho_mollo_milliere_2026_vector_grounding_problem} build
the possibility of misrepresentation into their account, and
\textcite{attah_2025_do_language_models_lack_communicative_intentions}
discusses conversational repair. Their treatments stop short of the live routes from
discrepancy to revision at issue here, and neither is committed to denying that such routes
matter.} Establishing content and establishing the capacity to detect and repair fresh
errors are different achievements, and the second doesn't follow from the first.

Distinguishing kinds of grounding still leaves open which differences in epistemic
architecture explain patterned truth-relevant success and failure. Truth-relevant success
depends on \term{stabilizing routes}\footnote{In
\posscite{grindrod_2024_large_language_models_linguistic_intentionality} reading of
Millikan, a \term{stabilizing function} is instead conventional and inter-agent: it sits
between speaker and hearer. It explains why a linguistic device keeps being reproduced,
and he puts the notion to work on language models. A device can have a robust stabilizing
function in Millikan's sense while preserving only the effects of the routes at issue
here. That's arguably what occurs in the text-only case: training preserves effects of
earlier coordination without retaining the coordinating routes themselves.} such as perception, testimony and records,
coherence constraints, measurement, intervention, checking and revision, and feedback
from practical success or failure. Some bring target
information into production; others integrate, check, or correct what's already there.

An \term{arrangement} is a language model together with whatever prompts, records, tools,
sensors, checkers, or reviewers are live for a particular target and task.\footnote{This
explanatory unit echoes the extended-mind treatment of functionally integrated external
resources as parts of a cognitive system \citep{clark_chalmers_1998_extended_mind}. It needn't
itself be a cognitive subject.} Its
\term{profile} records which routes constrain it and how those routes are coupled: which
can check, calibrate, or correct which others.

An output is \term{answerable} to a target to the extent that discrepancies can affect what
the arrangement produces or accepts, including through later correction or withdrawal.
That influence may be inherited, delayed, or supplied by external audit. The arrangement
has \term{corrective control} only when live routes let it detect and repair a fresh
discrepancy.

The arrangement, not the language model or its output alone, bears the profile and any
corrective control. Nor can it include the surrounding practice without limit: doing so
would credit every deployment with the epistemic resources of human science. Its boundary
is causal. Information outside it can't correct the output. The decomposition is an account
of epistemic architecture for explaining and predicting performance and guiding
intervention, not a semantic theory or scalar metric. It asks why grounding leaves that
boundary untouched and whether distinguishing inherited constraint from live correction
improves prediction or intervention.

\begin{figure}[H]
\centering
\begin{tikzpicture}[node distance=0pt]
  \node[anchor=west, font=\small\bfseries] at (-0.35,4.95)
        {(a) Historical inheritance};
  \node[font=\scriptsize\itshape, align=center, text width=10cm] at (6.15,4.42)
        {Derivative answerability only if target-sensitive correction survives selection and training.};

  \node[ttpfigbox, text width=1.55cm] (past) at (0.55,3.4)
        {past target\\states};
  \node[ttpfigbox, text width=2.2cm] (practice) at (3.1,3.4)
        {checking and correction\\in source practices};
  \node[ttpfigbox, text width=1.9cm] (corpus) at (5.95,3.4)
        {text selected\\for training};
  \node[ttpfigbox, text width=1.85cm] (modelstate) at (8.65,3.4)
        {trained\\model state};
  \node[ttpfigbox, text width=1.65cm] (candidate-a) at (11.35,3.4)
        {current\\output};

  \draw[ttpar] (past) -- (practice);
  \draw[ttpar] (practice) -- (corpus);
  \draw[ttpar] (corpus) -- (modelstate);
  \draw[ttpar] (modelstate) -- (candidate-a);

  \node[ttpfigbox, text width=1.85cm] (current-change) at (0.7,1.95)
        {target-relevant change\\after training};
  \node[font=\Large] (blocked-a) at (10.2,2.72) {$\times$};
  \draw[ttpblocked] (current-change.east) --
        node[pos=0.56, below, font=\scriptsize\itshape]
        {no live path once training is fixed}
        (blocked-a.west);

  \node[anchor=west, font=\small\bfseries] at (-0.35,0.75)
        {(b) Live evidence and correction};

  \node[ttpfigbox, text width=1.9cm] (current-target) at (0.7,-0.35)
        {current target\\state or change};
  \node[ttpfigbox, text width=2.15cm] (evidence) at (3.75,-0.35)
        {live evidence:\\record, sensor, or\\test result};
  \node[ttpfigbox, text width=1.65cm] (candidate-b) at (6.75,-0.35)
        {candidate\\output};
  \node[ttpfigbox, text width=2.55cm] (checking) at (6.75,-2.25)
        {compare output with evidence;\\detect and attribute\\a discrepancy};
  \node[ttpfigbox, text width=1.85cm] (uptake) at (9.55,-2.25)
        {act on discrepancy\\and revise};
  \node[ttpfigbox, text width=1.75cm] (revised) at (12.0,-2.25)
        {revised\\output};

  \draw[ttpar] (current-target) -- (evidence);
  \draw[ttpar] (evidence) -- (candidate-b);
  \draw[ttpar] (evidence.south) to[out=-70,in=165] (checking.west);
  \draw[ttpar] (candidate-b) -- (checking);
  \draw[ttpar] (checking) -- (uptake);
  \draw[ttpar] (uptake) -- (revised);
\end{tikzpicture}
\caption{Historical inheritance versus live correction. Solid arrows show possible causal
influence, not answerability or warrant; the dashed line ending at $\times$ shows no
directed path. In panel (a), target-sensitive correction in source practices can survive
selection and training, supplying derivative answerability, although a fresh target-relevant
change can't reach current output once training is fixed. In panel (b), live evidence
constrains a candidate, enabling detection, attribution, and revision. This directed path
supplies live answerability; it supplies corrective control only when evidence reaches the
arrangement, checking is sufficiently independent, the discrepancy is attributable, and
revision improves the output. Section~\ref{sec:projectibility}
separates those conditions into five graded features.}\label{fig:routes}
\end{figure}
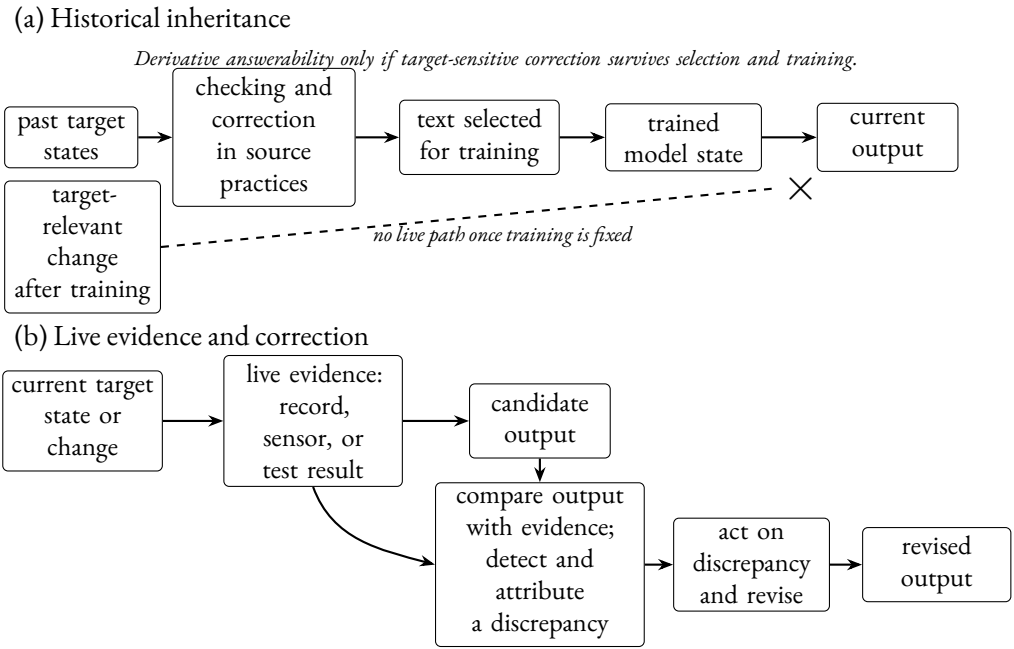

LLMs matter because they separate routes that human epistemic life normally bundles
together. The separation is clearest in the limiting case of an arrangement whose contact with
the world remains mediated mainly by previously produced human text rather than live
retrieval, perceptual inputs, execution results, or action outcomes. That case supplies a
baseline, not a description of every LLM arrangement. Where training preserves the effects
of target-sensitive correction in source practices, it can supply
\term{derivative answerability}. \term{Live answerability} adds a temporal condition: a
fresh discrepancy has to be able to reach current production. Relative to the baseline,
retrieval, tools, code execution, multimodal input, and action-guided feedback add live
routes, reducing reliance on inherited constraints for the tasks those routes can serve.
Their diagnostic value depends on which route each adds and where any resulting improvement
transfers.
Figure~\ref{fig:routes} sets out the difference.

An intervention should help only where it supplies a route the task needs. Self-consistency
should help most where the missing support is coherence; retrieval should help most where
the missing support is record access.
Feedback based on human preference judgments should improve conversational fit before it
improves answerability.
Multimodal input should help some perceptual tasks before it helps calibrated measurement.
These contrasts provide an initial test of the proposed distinctions among routes. Indiscriminate transfer would
count against the proposed decomposition.

These contrasts compare epistemic routes while leaving open what truth consists in.
Correspondence, coherence, pragmatism, reliabilism, and social epistemology each identify a
relation or process that can constrain representations in ways relevant to getting things
right. Deflationism leaves those
explanatory questions open while denying that truth itself requires a substantive
reduction. The comparison concerns how these relations and processes combine within an
arrangement and what follows when one is missing. It's a framework for comparison, not a
reconciliation of the theories.

This paper develops that account of epistemic architecture in three steps. First, it isolates the
live, task-indexed routes by which a fresh discrepancy reaches production as something to
be explained in its own right rather than a corollary of content. A positive grounding verdict can
settle what a system's representations mean or refer to without licensing an inference
about which fresh errors a deployed arrangement can catch.

Second, it separates the profile from corrective control. Route identity and coupling
determine the kind of target constraint available and where it should help. Five
arrangement-level features then diagnose whether target-relevant variation can expose a
sufficiently independent and attributable discrepancy and produce a target-improving
revision. They also separate the capacity to correct from observed correction performance.

Third, it turns the full decomposition into predictions that can fail. Route-by-task
interactions test whether the proposed routes track selective transfer. An information
account can predict them too by appealing to the task-relevant material available to the
arrangement. The stronger test is held-out transport to new cases. A decomposition
fixed in advance should predict new combinations of routes, tasks, and perturbations better
than a prespecified model of the task-relevant information supplied to an arrangement. It
should also guide intervention choice without conceptual refitting.

Sections~\ref{sec:target}--\ref{sec:projectibility} fix the target, explain route profiles,
and define the corrective-control features. Section~\ref{sec:llm} applies the decomposition
to language-model arrangements and derives the empirical transfer tests.
Section~\ref{sec:adjudication} compares it with component theories and addresses the
principal objections.

\transitionpagehead{1\quad Introduction: from grounding to corrective control}
\section{The target: truth-relevant representational success}\label{sec:target}

Before evidence about an arrangement's profile can support expectations about nearby cases,
its target has to be fixed. This paper doesn't profile
truth considered as an abstract property or as the role of the truth predicate, neither of
which has perceptual, testimonial, or instrumental mechanisms as parts. It profiles
\term{truth-relevant representational success}: beliefs, assertions, inscriptions,
measurements, models, and inferential outputs assessed as getting things right.

That target extends beyond central \term{truthbearers}. Beliefs and assertions can be true
or false in the ordinary sense. Neighbouring cases support truth-ascription by yielding,
recording, or constraining claims: a measurement, map, statistical model, or experimental
display may be accurate, calibrated, or well-fitted in ways that make associated
propositions true.

These cases often divide the representational labour. In a lead-exposure case, the assay
reading, calibration log, database entry, and clinician's assertion do different jobs.
Only the last may assert that the sample contains 3.2 mg of lead, but the others make that
assertion available, checkable, and defeasible. A map can likewise preserve spatial
relations that support claims about location, and a model can constrain which inferences
remain live. The object of analysis is the background through which such outputs support getting
things right.

Fixing the target this way leaves a reductive definition of truth to one side.
Deflationary and minimalist theories can give the truth predicate a thin logical or
generalizing role while keeping a substantive causal story out of the basic theory of
truth \citep{horwich_1998_truth}. That point can stand. What remains to explain is why
truth-relevant representational successes cluster and fail in patterned ways, and when evidence
about them supports expectations about nearby cases.

The target leaves correspondence intact. Correspondence theories preserve the idea
that truth involves answerability to how things are
\citep{david_1994_correspondence_disquotation,kirkham_1992_theories_truth}. Correspondence
fixes the target of answerability; it doesn't specify the supporting routes.

\term{Answerability} generalizes the constraint beyond central truthbearers. A
representation is answerable to a target when discrepancies between it and the target can
bear on its production, acceptance, correction, or withdrawal. Strictly, profiles belong to
arrangements, and an output is answerable relative to the arrangement that produces,
maintains, and revises it. For brevity, I sometimes call an output answerable without
restating the arrangement that supports the relation.\footnote{The
lexical history of \mention{true} is illustrative, not evidential. The \textit{Oxford English
Dictionary} records senses of loyalty, veracity, accordance with fact, accurate fit, and
mechanical alignment \citep{oed_true_2026}. They give the broader constraint relation
historical precedent, but inferring the analysis from them would be an etymological fallacy.}

Answerability is graded because routes differ. A calibrated instrument is answerable when
checks, recalibrations, and downstream failures can expose discrepancies. A coherent but
disconnected arrangement is less answerable when the relevant fact has no route into
production. Audit, replication, practical failure, or later correction can still connect a
fact to an output, even after delay. Specific, repeatable routes that readily expose the
relevant error make that connection stronger; generic or loosely targeted routes make it
weaker.

Different routes make the gradation concrete. Repeated weighing, standard weights, and
failed downstream uses can expose scale drift. Records, other witnesses, and later events
can defeat a mistaken report. A language model that generates a plausible citation from textual
pattern alone has so far been shaped by genre, with those checks absent from the production
route. Its citation may still be right, but its rightness is less supported by the route
that produced it.

Code execution supplies a contrasting live route. A language model can write a function, run it,
inspect a compiler error or failed test, and revise the code. A one-shot generator lacks
that route even when its first answer happens to work.

The contrast survives when content is held fixed. \mention{The sample contains 3.2 mg of
lead} is strongly answerable when it's generated by a calibrated assay whose logs,
standards, and downstream checks can expose error. The same string is weakly answerable
when a text-only arrangement produces it because such numbers fit the genre. The truth conditions
haven't changed; the route by which the claim became available has.

Pragmatist theories identify practical correction as another route to answerability.
Inquiry, action, and successful practice can correct locally coherent or socially inherited
representations with weak world-connection
\citep{james_1907_pragmatism,dewey_1941_propositions_warranted_assertibility_truth}. That
doesn't make truth identical with what works. Practical success helps arrangements become
answerable to their targets.

Here, \term{truth-tracking} has a modest sense. It names patterned support
for representational success, whereas
\posscite{nozick_1981_philosophical_explanations} truth-tracking account specifies
subjunctive conditions on knowledge. Patterned support lets us expect some representational
successes to survive checks, perturbations, and neighbouring applications better than
others.

\posscite{boyd_2019_rethinking_natural_kinds_reference_truth} accommodationism explains
why such support can be patterned. Reference and truth depend on more than definitions in
isolation; they're achievements of practices that align perceptual, instrumental,
cognitive, and representational activity with relevant causal structures. That account
raises a narrower diagnostic question: how strongly do those routes and practices
constrain a given arrangement?

For LLMs, that diagnostic asks how much of this background a given arrangement inherits or
instantiates. A text-only arrangement may inherit patterns shaped by testimony, statistical
regularities, and local coherence constraints while lacking live perception, measurement,
intervention, and ordinary practical correction. That profile makes
truth-relevant assessment possible. The decomposition predicts uneven, empirically
diagnostic success.

With the target now fixed, which route patterns support projection to nearby cases?

\transitionpagehead{2\quad The target: truth-relevant representational success}
\section{Routes and profiles}\label{sec:profile}

\posscite{Goodman1955} new riddle asks which predicates support \term{projection} from
known to unknown cases. Here, \term{projectibility} concerns whether evidence about an
arrangement's route profile warrants expectations about its likely success or failure on
nearby tasks.

\posscite{khalidi2013} account provides the causal-network framework for this use of the profile. Natural
categories earn their status by supporting prediction and explanation, and they do so
when associated properties stand in causal relations that support projections from some
properties to others.

Applied here, Khalidi's framework distinguishes the evidence that warrants a projection
from the world-side relations that make it reliable. Relevant routes include perception, testimony and records,
coherence constraints, measurement, intervention, checking and revision, and feedback
from practical success or failure. Together, these routes make representational
outputs more or less answerable to the world.
Khalidi's later formulation of natural kinds as \enquote{nodes in causal networks}
supplies the graph vocabulary \citep[1387]{khalidi_2018_natural_kinds_nodes_causal_networks}. On
his account a kind is a node, a highly connected vertex whose core properties give rise to
derivative ones, and the network is what the node sits in. A profile isn't a kind in that
sense. I take from him the discipline that connectivity is the proposed
world-side ground of the projections, not a claim that truth-tracking is a node with a core.

Routes do different jobs. Some bring target information into the arrangement, some
integrate it with existing representations, some expose discrepancies, and some return
practical failure to later production. A route can do more than one of these jobs. The
routes' corrective significance depends on whether, together, they let the arrangement
detect and repair a fresh discrepancy.

Calling the result a profile marks two points. First, different routes can dominate in
different domains. Proof norms matter more in mathematics than ordinary perception does;
calibration matters more in laboratory measurement than conversational fluency does.
Second, the epistemic significance often lies in the relations among routes. Testimony matters
differently when it's backed by records, instruments, and correction than when it's an
isolated report.

Part of this decomposition comes from Boyd's property-cluster account
\citep{boyd1991,boyd1999}. Causally related property clusters can make real kinds support
projection without essences.
\posscite{boyd_2019_rethinking_natural_kinds_reference_truth} accommodationism explains
why such support can be patterned. Reference and truth are achievements of practices that
align perceptual, instrumental, cognitive, and representational activity with relevant
causal structures, not products of definitions in isolation. The narrower question here is
how strongly those practices constrain a given arrangement. Their routes form a coupled
pattern, not coordinate epistemic goods.

That accommodation is a gradual fitting of representational practice to the
structures it's used to track. Practices of using scientific terms, producing instrument
readings, and applying ordinary object names become more reliable as perception,
measurement, inference, correction, and use are adjusted together. Definitions alone don't
secure the fit; practices that make error detectable and revision possible maintain it.

Consider \mention{fever}. Feeling hot, thermometer readings,
calibration standards, thresholds, records, and revised estimates of normal body
temperature have adjusted one another over time. The term remains useful because these
routes expose error and let the practice revise its measurements.

Applied to text-trained arrangements, the analysis remains continuous with Boyd at the
explanatory level and diagnostic at the arrangement level. Accommodationism tells us why
representational practices can be answerable to causal structure; the profile records how
strongly the relevant routes constrain a given arrangement. A text-trained arrangement can
inherit stable linguistic and testimonial patterns while its
instrumental, perceptual, and practical routes remain attenuated.

Everyday judgment and coordination draw on perception, action, memory, testimony, and
practical failure. These are the routes on which error has imposed costs longest, and they
have repeatedly corrected one another. A language-model arrangement that lacks perceptual
and verification routes should show a corresponding failure pattern.

In arrangements for frontier science and remote historical fact, the routes may instead be
sparsely linked. Error cost is delayed, diffuse, or beyond current instruments. There may be
inference norms, expert testimony, and coherence constraints, but fewer shared routes of
correction. Shared selection or training history, accumulated correction procedures, and
measurement infrastructure are often limited there too.

Network boundaries still depend on the chosen level of abstraction.
\textcite{onishi_serpico_2022} show this for causal graphs, and \textcite{craver_2009} makes
the same point for mechanism talk. Following \posscite{khalidi2013} distinction between
epistemic purposes and non-epistemic practical or aesthetic ones, the purpose here is to
represent a target domain well enough for relevant inferential, practical, or corrective
success under perturbation. Connections can be sparse without dissolving an arrangement's
profile relative to that purpose.

\transitionpagehead{3\quad Routes and profiles}
\section{Corrective control and bounded projections}\label{sec:projectibility}

Routes and their relations define the profile, but they don't by themselves warrant
projections about further truth-relevant success or failure.

Perturbation makes the declared purpose empirically testable. It distinguishes lucky fit from
robust answerability. A claim based only on success in the training cases, the familiar
environment, or the current conversational context supports only narrow projections. It earns stronger
warrant when the representation continues to guide action, inference, measurement, and
correction after conditions change.

Testing that persistence requires only a modest causal framework.
\textcite{pearl2009,pearl2010foundations}
distinguishes association, intervention, and counterfactuals. Observed route identities and
relations describe the initial profile. Perturbation helps identify mechanism, and projectibility
asks counterfactually what would hold nearby. \textcite{woodward2003MakingThingsHappen,woodward2001}
supplies the manipulability reading: a route is explanatory when interventions on it
change performance in stable, predictable ways. The argument doesn't require a fully
specified structural causal model.

What matters is invariance under the declared changes. A route
pattern can support bounded expectations without being actively maintained by correction.
Corrective control is one way such robustness can be produced, but neither past stability
nor projectibility by itself establishes live error detection and repair%
\ifttpanonymous
.
\else
~\citep{reynolds2026projectibilityHistory}.
\fi

\term{Corrective control} adds capacities for discrepancy detection and target-improving
change. Corrective control is relative to a target and perturbation range.
\mention{Target-sensitive} is neutral about the target's epistemic value: an
arrangement can track rater approval while its assertions become less accurate.
\mention{Target-improving} means closer to the declared target; it is truth-improving only
when that target bears appropriately on getting things right. Within the declared bounds,
corrective control depends on five features of the arrangement's architecture:
\begin{enumerate}[label=\arabic*.,ref=\arabic*]
    \item \label{feat:target-access}\term{Target access}: target-relevant variation reaches
    the arrangement through an operative route.
    \item \label{feat:detectability}\term{Detectability}: a discrepancy can be registered
    by that arrangement against a standard it can consult.
    \item \label{feat:checking-independence}\term{Checking-route independence}: the checking
    signal doesn't merely reproduce the production route's error.
    \item \label{feat:attribution}\term{Attribution}: the discrepancy is specific enough to
    guide a response.
    \item \label{feat:effective-uptake}\term{Effective uptake}: the capacity to make a
    target-improving change early enough when a relevant discrepancy occurs, is detected,
    and is attributable.
\end{enumerate}

For the live path in Figure~\ref{fig:routes}(b), each feature is necessary at least
minimally. If target variation can't arrive, register, supply non-redundant evidence, guide
a response, or alter production in a target-improving direction, that path can't correct.
The features also have to be jointly instantiated along a connected correction path; capacities
scattered across disconnected routes don't combine into corrective control.
Above that minimum, the features are graded and potentially dependent; the numbering is for
reference, not rank. Target access, for example, belongs to the arrangement rather than to a sensor,
record, or checker considered alone: a route may be present without transmitting the
relevant variation for the declared task. Likewise, a checker's errors may partly correlate
with the generator's, attribution may be more or less precise, and uptake may be delayed or
probabilistic.

A single architectural change can improve several features. Giving a checker access to
repository state, for example, can improve target access and reduce its error correlation
with the generator. This account ranks one arrangement above another when it instantiates
every feature at least as fully and some more fully. Trade-offs needn't be rankable: greater
checking-route independence can come with weaker attribution. Nothing here licenses a
weighted score across the five.

These five features characterize corrective capacity, not observed performance. Capacity
concerns what the architecture can do. Exercise frequency records how often a route
actually alters production. \term{Conditional repair performance} is the probability of an
appropriate target-improving response when a relevant discrepancy is present, considered
alongside the rate of unnecessary or harmful intervention.

The target, perturbation, and criterion for improvement are declared first rather than
inferred from favourable outcomes; conditional repair and false-intervention rates then
test whether the arrangement realizes the capacity.

Effective uptake is a capacity, not an exercise rate. A near-flawless arrangement in a
benign environment may rarely need to exercise it; a poor generator may trigger a mediocre
checker often. Nor does every alteration count as repair, since a change can move the
output further from the target. Effective uptake asks whether a detected,
attributable discrepancy can produce a timely, target-improving change. Detection,
exercise, and uptake after detection can be reported as component rates. Audited
end-to-end correction and false-intervention rates provide evidence about capacity and
observed performance. Raw repair frequency isn't what corrective control consists in.

Among the five, checking-route independence alone compares one route's errors with
another's. No task-free statistic defines it. In the proposed checker test, the operative
quantity is conditional error overlap: how often the checker repeats the generator's error
when the generator is wrong, considered alongside false interventions on correct outputs.
Lineage and marginal error correlation can bear on that quantity but don't define it.
Target access, detectability, or effective uptake can be present with independence or
without it.

These conditions fix the relation to answerability rather than competing with it.
Answerability comes in degrees, and its strength depends on how specific, repeatable, and
proximate the route is. Corrective control is stronger than answerability alone: it
requires both discrepancy detection and an effective route to repair. So an arrangement
can be answerable in some measure while lacking corrective control over the targets at
issue.

A thermometer connected to an independently calibrated drift monitor illustrates the
distinction. Suppose it consults a separate standard, raises a correct
alert, and every alert is ignored. Target access, detectability, checking-route
independence, and attribution are present. Effective uptake is absent: no route connects
the alert to what the arrangement does. Because the people ignoring it are part of the
arrangement, that's a missing capacity rather than a low correction rate.

Route structure also supports a bounded projection. A value produced through a live assay,
calibration log, and error-reporting practice warrants stronger expectations about nearby
cases than one copied from yesterday's database, because changes in the sample or
instrument can affect the former. The range of warranted expectations still depends on the
declared target, task, and perturbation.

\transitionpagehead{4\quad Corrective control and bounded projections}
\section[Language-model arrangements]{Language-model arrangements: inherited structure and live
routes}\label{sec:llm}

\subsection{The arrangement and its routes}\label{ssec:arrangement}

The five features now provide a way to analyse the LLM pressure case. The first step is to
fix the output-producing arrangement for the target and task. It comprises the language
model and prompt, together with any retrieved records, tools, sensors, checkers, human
reviewers, or update routes that are live.

Multimodal foundation models are included when their linguistic outputs are under
assessment. Multimodal training can supply inherited perceptual structure; a runtime
image, audio, video, or sensor input adds a live route only when it's available and
integrated for the task.

Retrieval belongs to the arrangement when it runs, not to the language model. Human appropriation changes the arrangement
rather than adding a gloss to a fixed one.\footnote{Preparation of this paper supplies
mundane examples. When a coding agent reported sources unavailable, directing it to a
local reference library added a live source route. When criticism was retained in the
conversation and the author required revision, human review became a live correction
route. Neither belonged to the language model alone.} The effects of training history belong to the
arrangement's causal history.

The application separates inherited constraints from live routes. Training can shape
production without giving the arrangement
present access to the source practice. A live route can transmit a new record,
measurement, or check, and it may also support correction. A purported route that supplies
no target-relevant constraint is decorative: the signs of experiment, source use,
or measurement remain while the constraint tends to zero. The comparison is
interventionist: changing the active routes should change the expected pattern of success
and failure.

Live correction also has a horizon. In a stateless chat, a correct, attributable
counterexample that exposes a relevant discrepancy and causes a revision supplies
effective uptake within the session and nothing beyond it. The weights don't move, the
next session starts where the last one began, and no other user benefits.
The arrangement has corrective control for one conversation, not for the policy deployed
in later conversations.\footnote{External state can extend this horizon without changing
model weights. Claude Code's project memory and OpenAI's persistent conversation state
become live routes when later sessions load them and they affect production
\citep{anthropic2026claudecodememory,openai2026conversationstate}.}

Within each profile, the five features from Section~\ref{sec:projectibility} determine
whether the live routes supply corrective control.

Training can supply derivative answerability, but only on a condition. Discrepancies
already registered in the source practices have to shape present production through the
training history. Causal descent isn't enough by itself. Dense propaganda
can leave distributional traces as persistent as those of a measurement literature. A
language model trained on it can inherit the shape of a practice whose strongest corrective pressures track
audience response, party discipline, or persuasive success rather than the empirical
target at issue. What has to survive the training history is the effect of corrections
made against that target rather than against audience response, rater approval, or a party line.

Three things come apart here. Causal ancestry requires only that material included
in training shaped present production. \term{Derivative answerability} requires the
inherited effects of target-sensitive correction. \term{Live answerability} requires a
current route from a fresh discrepancy to production (Figure~\ref{fig:routes}).

\textcite{lederman_mahowald_2024_bibliotechnism} provide a nearby account of inherited
meaning. Their bibliotechnism allows novel generated text to be derivatively meaningful
when its production remains causally sensitive to the intelligibility of human-produced
training text. Derivative answerability requires more: the inherited constraint has to
preserve the effects of target-sensitive correction, not intelligibility alone.

Derivative answerability by itself never supplies live answerability. Once the training history is
fixed, a discrepancy arising after it has no way to reach production. Training against
verifiable rewards tests this boundary. When a language model is optimized against unit
tests, a proof checker, or execution results, a discrepancy between output and an
independently specified target is detected and does alter production. All five features can
be instantiated while the verifier runs, to the extent that it's adequate to the target,
sufficiently independent of the generator, and attributable in its signal. Tests can be
incomplete, attribution coarse, and verifier errors correlated with the generator's.

The training loop has then exercised corrective control over model outputs; the deployed
arrangement isn't thereby correctable. The loop supplies that control only over the targets its
verifier covers and only while it runs. A language model that has
learned arithmetic against a checker still can't tell you whether today's figure is right.
Verifiable rewards can strengthen derivative answerability, since the corrections they
encode were made against an independently specified target rather than against approval alone.
They don't make it live.

A live route can supply target access. It supports corrective control only if the
arrangement can detect and attribute a discrepancy, receive a sufficiently independent
signal, and act on it. A merely decorative invocation resembles such a route while
supplying none of these features. Checking-route independence depends on the relation
between routes, not on one route in isolation.

Corrective control doesn't extend beyond the arrangement's live routes, and it's bounded
target by target. A user who checks a claim can supply corrective capacity for that claim.
A later reader is part of the relevant arrangement only when the check can still affect the
output's acceptance, correction, withdrawal, or downstream use within the declared horizon.
Checking frequency fixes the route's coverage across claims; appropriate
improvement when a relevant discrepancy is present, considered with harmful intervention,
fixes conditional repair performance. The ordinary case isn't an arrangement with no
corrective control anywhere, but one
asked about targets that none of its live routes can independently check. That boundary
doesn't affect grounding, but corrective control extends no further.

\subsection{Route-specific interventions}

Retrieval augmentation queries an external corpus or search system and places retrieved material into the
language model's context; it adds record access when it works. Its characteristic failures are
diagnostic rather than incidental: a stale index supplies a record of the wrong time, a
chunk boundary supplies half a record, and an embedding match on wording rather than
content supplies a passage that resembles the answer. Each leaves the archival route
nominally present and materially absent: decorative retrieval implemented by a real
mechanism rather than produced by a turn of phrase.

Tool use routes a subtask through a calculator, database, code interpreter, application
programming interface, or instrument; it adds calculation or measurement only when the
right tool is called and its
result is integrated. When execution results guide revision, tool use adds a corrective
loop.

Self-consistency prompting samples multiple completions or reasoning paths and privileges
convergence among them; it strengthens internal convergence. An external check requires a
route to the relevant facts whose signal isn't generated by the same convergence process.

Instruction tuning and reinforcement learning from human feedback (RLHF) change a language
model's policy through demonstrations, preferences, benchmarks, or evaluations; they add a social or evaluative correction
route whose target depends on the feedback.\footnote{Preference signals have factual
targets. Rater approval of affirmation, deference, or confident presentation is a fact
about a population and interaction setting, and RLHF can make the resulting arrangement's behaviour
strongly answerable to that fact. The failure mode is target mismatch: the policy can
track the rater distribution accurately while its assertions are assessed against a
different, object-level target. Sycophancy is the clean case, where an arrangement accurately
tracks the wrong target.}

Multimodal input supplies image, audio, video, or sensor information; by itself, it adds
perceptual input. Calibrated measurement is a separate route.

Multimodality matters because it seems to answer
\posscite{harnad_1990_symbol_grounding_problem} sensorimotor prescription. A finer-grained
diagnosis separates perception from action, correction, and use. An arrangement built
around a vision-language model trained on image-caption corpora
inherits perceptual structure only through selected, framed, and captioned images. A
deployment-time image input adds a live perceptual route; calibrated measurement,
intervention, and action-guided correction remain separate routes.

Multimodal input should improve performance unevenly. It should reduce some failures in
object classification and leave measurement- and intervention-dependent tasks fragile.
It should also leave room for object hallucination, where generated descriptions include
objects not present in the image
\citep{li_etal_2023_evaluating_object_hallucination_lvlms}. That pattern is what this
account expects when target access is strengthened and the correction routes are left
as they were.

These interventions can be combined, so the profiles summarized below are neither
mutually exclusive product classes nor a scale. The text-only row remains a
task-relative limiting case. Even that limiting arrangement inherits reports written by perceivers, explanations corrected by
teachers, institutional records, and descriptions shaped by instruments. It can
support genuine epistemic achievements while occupying a different role from a speaker or
hearer in a testimonial exchange.

Table~\ref{tab:llm-profiles} summarizes the architecture-sensitive predictions.

\begin{table}[!t]
\centering
\small
\begin{tabular}{@{}>{\raggedright\arraybackslash}p{0.20\linewidth}>{\raggedright\arraybackslash}p{0.28\linewidth}>{\raggedright\arraybackslash}p{0.42\linewidth}@{}}
\toprule
Arrangement type & Principal architectural resources & Directional prediction \\
\midrule
Text-only arrangement with a pretrained language model & Training-distribution regularities and local coherence & Strong where dense,
stable text is a good proxy; fragile for recent, obscure, perceptual, or measurement-dependent facts. \\
Instruction-tuned or RLHF arrangement & Rater feedback, demonstrations, and benchmark evaluation & Better conversational fit; truth-relevant gains should concentrate where raters or benchmarks have target-sensitive evidence. Independently informative checking routes still have to be supplied. \\
Retrieval-augmented arrangement & Live access to records and institutional sources & Better on updated or source-backed recall; remaining fragility where the missing route is experiment, perception, or intervention. \\
Tool-using or coding arrangement & Calculators, databases, code execution, instruments, or application programming interfaces & Better where the right tool is called and interpreted; strongest when runtime or test results guide revision; new failures from tool choice, stale data, and test-as-proxy effects. \\
Multimodal arrangement & Perceptual input in image, audio, video, or sensor form & Better on perceptually anchored tasks when runtime input bears on the target; no general gain on calibrated measurement or practical correction. \\
Action-guided or expert-reviewed arrangement & Practical feedback, review, and downstream failure signals & Stronger corrective capacity where feedback is timely and domain-competent; weaker where review is sparse or proxy-based. \\
\bottomrule
\end{tabular}
\caption{Architecture-sensitive predictions. Entries summarize expected directions; the
testable claims are the prespecified route-by-task interactions.}
\label{tab:llm-profiles}
\end{table}

\subsection{Inherited structure and missing anchors}\label{ssec:inheritance}

Textual inheritance explains both the competence and the fragility of text-only arrangements.
Human text is saturated with world-directed practices of looking, measuring, arguing,
repairing, citing, and correcting. The deployed arrangement usually remains outside those
practices when it produces a fresh answer. It draws on their products with attenuated access to
their present checks.

Text-only training can still produce more structure than a bare \mention{textual
residue} slogan suggests. The generative pretrained transformer model Othello-GPT is a
challenge case: a sequence model trained to predict legal moves in a board game
developed an internal representation of board state, and interventions on that
representation affected its outputs \citep{li_etal_2023_emergent_world_representations}.
The result shows that inherited constraint can be strong when the target state is densely and
systematically encoded in the sequence.

Othello marks the strongest case for inherited structure: sequence training can recover a
latent target state when the record encodes it completely. Its broader reach is limited.
The game is finite, fully observable, and rule-governed; its legal moves are a deterministic
function of a state recoverable from the move sequence. The case identifies the
boundary of the claim rather than supplying evidence about domains whose relevant state is
absent from the record.

Internal world-model evidence should be strongest where the relevant
state variables are encoded in the sequence and the loss function penalizes violations
of their structure. It should weaken when success depends on variables absent from the
sequence, live measurement, changing local conditions, or intervention in the target
system. Successful transfer across those boundaries would count against this diagnosis.

Coherence also constrains these arrangements. Training and decoding reward genre fit,
local well-formedness, smooth inference, and continuation from context, supporting the
formal linguistic competence distinguished from functional world use by
\textcite{mahowald2024}. The same pressures can produce a plausible lab result, legal
summary, observation, or citation without a live experiment, court record, observation,
or catalogue entry.

That preservation condition predicts that textual inheritance should preserve
distributional patterns more consistently than archival
particulars. Particular strings, names, and sources can survive, so the claim isn't that
token-bound information never does. Training supplies no general relation of record
identity, provenance, and current checkability. A text-only language model can encode
statistical regularities in testimonial practice.
The resulting arrangement can handle a domain whose records are dense and stable while
fabricating the verbatim anchors (quotations, identifiers, author lists) on which the
practice's correction routines depend. The same distinction predicts that these failures should be especially
retrieval-sensitive: retrieval adds the archival route that distributional inheritance
lacks.

Hallucination plays a more specific role here. The LLM literature treats it as heterogeneous
\citep{huang_etal_2025_survey_hallucination_llms}. The narrower claim is that
text-only arrangements should be especially vulnerable to fluent, coherent, genre-appropriate
outputs that fail when the task requires a fresh or independently informative route to the facts.
Those failures should be common for obscure sources, recent events, local observations,
novel measurements, fine-grained quantities, and action-sensitive causal claims. They
should be rarer where dense, stable, redundant text is a good proxy.

Frankfurtian accounts of bullshit \citep{frankfurt_1986_on_bullshit} diagnose LLM output
differently. \textcite{hicks_humphries_slater_2024_chatgpt_bullshit} apply that diagnosis
to ChatGPT-style systems, arguing that their outputs are produced with indifference to
truth. On the Frankfurtian diagnosis, the system doesn't care where the facts are; on this
diagnosis, the relevant facts have no route into production or correction. That
missing route yields comparative predictions across retrieval, tools, multimodality, and
action-guided feedback. This diagnosis asks for a route, not a better attitude.

Mitigation results should be attributed to the route manipulated. Evidence that
retrieval reduces hallucination in conversation bears on record access, not on perception,
experiment, or intervention \citep{shuster2021retrieval}.

Action-guided arrangements expand corrective control when feedback is timely, attributable,
and tied to the failure. Section~\ref{sec:target} gives the code-execution case. A robot that
breaks a glass, misses a target, or fails a laboratory protocol can receive similarly direct
correction; delayed approval, engagement metrics, or sparse user edits give a chatbot a
slower, less attributable route. The label \mention{feedback} hides the difference unless
the route is specified.

\subsection{What the correction routes correct}

Instruction tuning, RLHF, benchmark evaluation, red-teaming, and expert review function as
correction routes, and they should be classified by what they correct. Their immediate
targets often include helpfulness, harmlessness, preference satisfaction, conversational
fit, and benchmark success. Those constraints can reduce some errors and can also reward
answer-shaped behaviour when users or raters prefer confidence, agreement, or plausible
explanation.

The verifier case also separates two forms of post-training. Rewards based on a rater's
approval target that approval; verifier-based rewards target an independently specified
criterion, subject to the verifier's adequacy.
The label \mention{post-training} covers both, but they build different profiles. Treating checker-based
reward as mere rater approval understates what it supplies; treating rater approval as
truth-directed feedback overstates it. For this analysis, the targets against which rewards
were computed matter more than the amount of post-training.

Work on sycophancy shows the risk of treating feedback based on human preference
judgments as truth-directed on its own \citep{shapira2026rlhf}. These processes strengthen
some social and evaluative constraints while leaving routes through perception,
measurement, and intervention indirect or absent.

Accounts of fine-tuning disagree because they seek to explain different things.
\textcite{grindrod_2024_large_language_models_linguistic_intentionality} treats it as a
relatively marginal procedure that can't be what makes the difference between
intentionality and its absence, since language models produce apparently meaningful text without
it. \textcite{ruyant_2026_what_do_llms_represent} treats it as constitutive of what the
system represents, because the norms of appropriateness a language model is tuned toward are what
fix its target.

Fine-tuning can be constitutive of appropriateness for a purpose, the phenomenon Ruyant
seeks to explain, because norm setters fix that purpose. It can remain
marginal to whether outputs are meaningful in Grindrod's sense. Neither verdict settles
whether fine-tuning provides target-directed correction. For rater feedback to provide such
correction, the judgment must draw on a target-sensitive route, such as checking against
records, tool results, or action outcomes; approval alone isn't enough. Treating all three questions as one quantity called
\mention{grounding}
makes the disagreement look unresolvable.

\subsection{What grounding leaves open}

Referential and teleosemantic arguments can establish genuine, though limited, content in
LLM states \citep{coelho_mollo_milliere_2026_vector_grounding_problem,
mallory_2026_teleosemantics_word_embeddings}. That conclusion isn't in dispute. The
remaining issue is whether the deployed arrangement has the live correction, measurement,
intervention, perception, or feedback routes required by the task.

\textcite{mandelkern_linzen_2024_do_language_models_words_refer} illustrate the dissociation. They
close with Izzy, isolated from
birth in a sensory-deprivation chamber and reaching the world only through a text screen,
and they argue that externalism dissolves the puzzle about how his words
refer. Izzy's
answerability is nonetheless confined to the testimonial and inferential routes his screen
supplies, with perception, measurement, and intervention absent. Secure reference
alongside a restricted profile is the dissociation at issue.

\textcite{pepp_2025_reference_without_intentions_llms} raises a stronger version. She
suggests that such systems may contact objects
through their exchanges with users, coming into contact with whatever shaped a user's
input. The inherited--live distinction splits that case. A live image-input route supplies
perceptual access. Contact with the objects that shaped a user's
text is inherited: the perceiving
was done by someone else, and nothing in the exchange reports back when the user was
wrong.

Bender and Koller's octopus and Mandelkern and Linzen's Izzy support opposing verdicts on
reference. Their shared limit is architectural. In each case, linguistic
traffic is the principal route, so a discrepancy that never enters that traffic can't
alter present production. What isn't on the wire can't correct what is.

\subsection{A deployed checking route}\label{ssec:deployment}

A deployment case illustrates detectability and effective uptake without providing
evidence of truth-directed correction. \textcite{anthropic2026automode} describes a coding
agent that routes each proposed action through a classifier that blocks operations it judges
irreversible, destructive, or aimed outside the working environment. The target is action
safety rather than factual accuracy. Nevertheless, the architecture instantiates the live
correction path in Figure~\ref{fig:routes}: a separate discrepancy signal becomes
behaviourally effective.

When its policy judgment is correct, the separate classifier can supply detection and
effective gating where the acting language model registers nothing itself. Adding it can
give the arrangement detectability and effective uptake through a second route rather than
through introspection.

A checker needn't be artificial. A human reviewer belongs to the same arrangement when the
reviewer receives the proposed action, has relevant evidence, and can block or revise it.

Separation alone doesn't secure checking-route independence. An external checker can
reproduce the generator's errors, while the generator given independently sourced
environmental information may not. Relevant dimensions include development provenance,
the environmental information available to the checker, objectives, and conditional error
structure. The reported hardening gave the classifier more environmental information,
including repository visibility and git state. Whether that change decorrelated its errors
wasn't measured, and lineage alone wouldn't settle the question. Measuring it requires
conditional error rates for checker and generator on a shared held-out set.

\subsection{Empirical transfer}\label{ssec:transfer}

The deployment case illustrates detectability and effective uptake while leaving
checking-route independence unresolved. The component tests below are validation arguments,
not attempts to read architecture directly off performance: each links an observed contrast
to a claim about route structure under declared auxiliary assumptions and manipulation
checks. The broader empirical question is whether route-specific interventions transfer as
the decomposition predicts. Improvements in one route should remain limited where a task
depends on a different route. If perception,
measurement, retrieval, tool use, expert correction, or action feedback is confirmed to be
active and reaches production but fails to alter the expected contrast, the proposed
decomposition has misdescribed the mechanisms. If text-only arrangements remain mainly
dependent on inherited textual accommodation and local coherence, the same vulnerability
should persist even as fluency improves.

A clinical design applies both retrieval and a measurement tool to two sets of questions.
Record-dependent questions ask for recommendations recoverable from guidelines or
institutional records. Patient-specific questions require a current observation,
measurement, calculation, or causal judgment about an intervention.
Retrieval should selectively help the record-dependent items, and a measurement tool the
measurement-dependent ones. The declared route-by-task interaction is the main test:
comparable gains from retrieval alone on both sets would count against the proposed route
distinction. These directional predictions require declared auxiliary assumptions about
task requirements, route invocation and integration, and the data distribution; the
decomposition alone doesn't fix an effect magnitude. Online Resource~1 specifies the item
construction, estimand, controls, manipulation checks, outcomes, and interpretation of
uncertainty.

This double dissociation tests selective transfer, but it isn't unique to the decomposition.
A relevance-sensitive information account can predict the same crossing by treating records
and measurements as different kinds of task-specific information. Assessing the strongest
information-based reply requires distinguishing three readings of coverage.

Input coverage concerns task-relevant content accessible at a declared architectural boundary
before the component under test produces its signal. The boundary and criteria for relevance
are fixed before outputs are observed. This restricted comparator allows two arrangements to
receive the same task information while differing in what they do with it.

Conditional coverage concerns what one route adds given what the other routes have
supplied. It can represent the greater value of a checker whose errors don't reproduce the
generator's. Effective coverage includes whatever information the arrangement can bring to
bear on the final output, given its provenance, checking relations, and uptake policy. It
summarizes the usable information supplied by the whole arrangement without decomposing how
it became usable.

Input coverage is the clearest competing account. Conditional coverage remains an empirical
comparator when fixed in advance, but it already incorporates checking-route independence and
joint error structure. An unrestricted effective-coverage measure recapitulates
nearly the whole route profile; without fixed
application conditions, it can fit any result post hoc by counting each helpful route as an
increase in coverage. It then summarizes the arrangement's usable information rather than
competing as a mechanism. Nothing here requires one description to serve both that summary
and route-specific intervention.

This yields an identification limit. For any mapping from the decomposition's variables and environments
to performance, an unrestricted coverage variable can be defined to reproduce that mapping.
Behavioural results alone can't identify which architecture produced the usable information.
This is a methodological limit, not a gap that a more ingenious behavioural contrast can
close. Once coverage is allowed to encode every route relation and uptake policy, effective
coverage is a reduced-form summary of usable information, while the decomposition is a
structural proposal about how that information becomes usable and where an intervention
should transfer.
Lineage, information flow, conditional error structure, and update routes supply the
additional evidence. The empirical comparison needs fixed application conditions specified
before the outcomes are observed.

The proposed decomposition faces the empirical burden. Route identity and coupling specify
the available target constraints; the five corrective features diagnose whether those
constraints can expose an independently informative discrepancy and produce a
target-improving response. Fix both components from architecture and history, then compare
the resulting model with prespecified models of input and conditional coverage. Its
predictors are route identities, task requirements, cross-route relations, the five features,
and declared interactions. Evaluate all three on held-out combinations without reclassifying
routes after seeing the outcomes. No weighted profile score is required. The projectibility
claim loses if either fixed coverage comparator predicts those cases equally well without
conceptual refitting.

The architectural counterexample in Figure~\ref{fig:routes} establishes the conceptual
point: grounding can remain fixed while a fresh discrepancy has no path to current
production. The decomposition makes a further empirical claim: its distinctions should
support prediction and intervention. Simulated data cannot establish that claim because
their generating process is stipulated. Online Resource~1 instead audits the proposed model
comparison before it is applied to real systems. Can the analysis recover a known
architectural advantage, recognize conditions under which a simpler model is sufficient,
and distinguish predictive improvement from better intervention choice?

World A is a positive control. Uptake is built into its generating process and omitted from
the comparator, so the result shows only that the analysis can recover that designed
difference. Worlds B--D test three ways the empirical claim can fail. Conditional coverage
is sufficient in B. In C, noisy measurements make the decomposition practically useless
even though route structure generates the outcomes. In D, it predicts repair better without
improving intervention choice.

A stronger model specified before the outcomes are observed should replace the present
comparator. If it performs as well with fewer commitments, the decomposition loses its
projective claim. A coverage model that includes attribution and uptake can reproduce World
A, but then it is an alternative parameterization of the same architecture rather than the
simpler rival. The exercise thus defines conditions under which the decomposition earns or
loses an empirical role.

For the practical claim, the decomposition has to improve diagnosis or the choice among
retrieval, measurement, independent checking, and enforced uptake
beyond whatever predictive gain it supplies. Evidence about how the routes have checked or
corrected one another (their \term{calibration history}) earns a place only where it helps
characterize the route relations, the five features, or their stability under perturbation.

A second component test varies checking-route independence
(feature~\ref{feat:checking-independence}). Compare a checker from the generator's lineage
with one trained and calibrated separately while matching supplied information, standalone
accuracy, and revision policy. The decomposition predicts greater corrective value from the
checker with lower conditional error overlap. An input-coverage account predicts no advantage
under the declared boundary; a conditional-coverage account can predict one by representing
the joint error structure that checking-route independence names.

Online Resource~1 also compares a frozen record with a live route to the same source under
update and no-update arms. The test operationalizes live answerability but doesn't
distinguish the decomposition from dynamic coverage, which predicts the same follow-up pattern.

Failure of a declared contrast, after confirming that the relevant route was invoked and
reached production in time, would count against the corresponding route distinction or
corrective feature. Repeated failure across prespecified component tests would defeat the
full decomposition.

\transitionpagehead{5\quad Language-model arrangements}
\section[Comparisons and objections]{What the decomposition adds and principal
objections}\label{sec:adjudication}

The language-model cases show what the decomposition adds to the component theories
canvassed earlier. Reliabilism foregrounds reliable processes
\citep{goldman_1979_what_is_justified_belief}; coherentism, integration constrained by
input \citep{bonjour_1985_structure_empirical_knowledge}; and pragmatism, correspondence,
and social epistemology supply other routes and target relations already introduced. The
comparison concerns how those routes interact within one arrangement and what should fail
when an interaction is removed.

\subsection{Reliabilism and relations among routes}

Relations among routes can change the evidential value of each. Perception checked by
testimony, testimony disciplined by records, and measurement corrected by intervention
have a calibration history. Evidence from that history matters only where it
predicts present error relations or stability under perturbation. The WHO Surgical Safety
Checklist gives the larger-scale pattern: copying the checklist without the team
coordination, local adaptation, coaching, and feedback involved in effective
implementation need not reproduce its effects
\citep{who_2009_surgical_safety_checklist,armstrong_etal_2022_surgical_safety_checklist}.

Reliability and answerability remain distinct. Reliability is statistical;
answerability is a counterfactual relation between a representation and its target,
relative to an arrangement; corrective control is an architectural capacity. An ordinary
accuracy record estimates the first, whereas audited correction and false-intervention
rates bear on the third.

The clinical contrast in Section~\ref{ssec:transfer} also shows why process individuation
matters. Describing a retrieval-backed arrangement as \mention{a system that answers
medical questions reliably} invites transfer from record-backed questions to tasks that
require live measurement, patient-specific observation, calculation, or intervention. A
careful reliabilist can instead individuate the process as retrieval-backed answering under
a specified profile of records, tools, and checks. That yields the same verdict.
Both views need an ex ante process description. The decomposition's distinctive
claim is that route identities and relations provide a principled way to fix it before
transfer is observed.

At arrangement scale, this raises reliabilism's generality problem
\citep{conee_feldman_1998_generality_problem_reliabilism}. Pretraining, filtering, RLHF,
retrieval, tools, prompts, and institutional uptake permit many process descriptions. If
selection history, correction procedures, measurement infrastructure, and current
architecture pick the description that supports projection, the account supplies the individuation
theory. The account loses that role if those variables don't improve held-out prediction or
intervention choice.

The projective basis is the arrangement's current route structure together with
whatever calibration history bears on its stability under perturbation. It isn't a single
process assessed apart from that background.

Route relations also matter when sources conflict. An inspector may judge by sight that a
platform is level while a calibrated inclinometer puts its slope outside tolerance and
technicians report that the instrument is functioning normally. No source has standing
priority; the comparison depends on assessed reliability and a history of mutual
correction. Agreement among sufficiently independent witnesses confirms more than
agreement inherited from a common source or error process
\citep{bovens_hartmann_2003_bayesian_epistemology,olsson_2005_against_coherence}.
Coherence faces the parallel limit: a false proposition can cohere with a specified set
\citep{russell_1907_nature_truth}, and corpus-grounded meaning can remain detached from a
live route by which target drift alters production
\citep{havlik_2024_meaning_understanding_llms_synthese}.

\subsection{Interest-relativity and genuine kinds}

One objection questions the profile's objectivity. Goodman's new riddle makes
projectibility depend on predicates embedded in inductive practice
\citep{Goodman1955}. Boyd makes the corresponding realism field-relative: kinds answer to
the accommodation demands of particular practical or scientific disciplines, and different
disciplines may require different carvings of the same domain \citep{boyd1991,boyd1999}.
Causal graphs and mechanism boundaries likewise depend on explanatory interests
\citep{craver_2009,onishi_serpico_2022}; no network fixes a unique, interest-free carving.
\textcite{khalidi2013} allows overlapping scientific domains to support different categories
and methods.

That relativity is compatible with genuine kinds. A field fixes which similarities, causal
relations, and perturbations matter; the world determines whether a classification supports
induction and explanation under those conditions. Here, the target, task, perturbation, and
failure condition are declared first. Interests select the projective claim, not its truth.
Route relations are then characterized before present performance through selection or
training history, accumulated correction procedures, and measurement infrastructure. Those
marks can establish occasions for one route to expose another's errors; longevity or
consensus alone can't.

Choosing the purpose after seeing the verdicts would make the account unfalsifiable.
Prospective characterization blocks that move. A failure on the declared held-out cases
requires the profile to be revised, narrowed prospectively, or abandoned rather than
rescued by redescribing the routes after the fact.

\subsection{Derivative testimony and fictional output}

Testimony ordinarily involves assertion, responsibility, trust, uptake, and defeaters
\citep{coady_1992_testimony,lackey_2008_learning_from_words,goldberg_2010_relying_on_others}.
Conversational systems strain such anthropocentric roles without simply becoming ordinary
testifiers \citep{freiman_2024_ai_testimony}. The claim here is derivative: a
text-trained language model inherits artifacts of testimonial practice, not testimonial
standing. It remains outside the current exchange and supplies neither a normal hearer who
trusts under standing norms nor a speaker who assumes responsibility.

That history carries epistemic support only when target-sensitive correction survives
selection and training. Section~\ref{ssec:inheritance} gives the resulting prediction:
distributional features of a practice should be preserved more consistently than source
identity, provenance, defeaters, or verbatim anchors. Causal descent without that
preservation supplies influence, not derivative answerability.

A related objection treats generated outputs as fiction. \textcite{ruyant_2026_what_do_llms_represent}
argues that a language model represents a constructed class of appropriate linguistic
production and that chatbot outputs are props in a game of make-believe. His own scope
condition is architectural: the generalization holds to the extent that direct worldly
inputs don't train the system. He also grants that generated text becomes dissertations,
emails, programs, news, or scientific prose when people appropriate it.

Appropriation needn't change the string's kind. The same string can be correct in the game,
false about the world, and effective as an artifact. What changes is the target and the
arrangement: measurement, review, and correction may become live when a person or
institution takes up the output. Target-indexed answerability extends the fictionalist
analysis rather than requiring its rejection.

\subsection{Scope}

The central case is empirical: it concerns targets whose relevant variation may reach an
arrangement through testimony and records, perception, measurement, or intervention.
Mathematics and logic provide a contrast. Proof and formal checking can supply correction
without empirical target access, showing that detection and correction needn't be
perceptual. Deflationists
may deny that truth has a substantive nature \citep{horwich_1998_truth}, while pluralists
distinguish ways of being true across domains. Neither view conflicts with the narrower
claim that evidence about different routes may warrant bounded projections among
truth-relevant representational successes. The account explains those projections, not
truth itself.

\transitionpagehead{6\quad Comparisons and objections}
\section{Conclusion}\label{sec:conclusion}

The octopus and Izzy support opposing verdicts on reference, but the central question
survives either one. Representations can have content, reference, and correctness
conditions even when the arrangement lacks live, sufficiently independent
routes from detected discrepancies to altered production. Grounding doesn't entail
corrective control.

Language-model arrangements expose the dissociation at scale. Their language models inherit
patterns shaped by testimony and coherence constraints. Those patterns supply derivative
answerability where source practices have already made target-sensitive corrections, while
the arrangements may lack live routes that detect a fresh error and carry it back into
production. Identifying which routes constrain an arrangement, how they're related, and
for which targets shows whether the missing corrective routes have been supplied or only
partly compensated for. Rich internal structure can arise where
the target domain is densely encoded in the sequence stream without supplying those live
routes elsewhere.

A thin truth predicate can coexist with thick truth-tracking practices. Recurrence of a
profile can support bounded expectations without live correction; corrective control adds
capacities for detecting and repairing fresh discrepancies. None of that architecture need
be built into truth itself.

The older epistemologies keep their subject matter. Each can represent relations among
the processes it describes, and reliabilists and social epistemologists routinely do. The
relevant routes normally operate jointly; language models make their separation visible.

The resulting framework compares arrangements rather than assigning systems to fixed
classes. Effective coverage can summarize whether an arrangement had usable information,
but it doesn't by itself say which intervention to choose or where a gain should transfer.

Retrieval, multimodal input, tool use, expert review, experimental embedding, and
action-guided feedback alter an arrangement's route profile. The empirical burden is to
specify in advance which route each intervention adds and where improvement should
transfer. The route-by-task interactions aren't unique to the decomposition. Its stronger
burden is to predict held-out combinations of interventions and tasks from a fixed decomposition
better than a prespecified input-coverage model. An arrangement can improve along one dimension
and remain fragile along another. Routes and coupling are characterized from causal and
architectural facts before performance is checked, and the predicted contrasts can fail.

Beyond LLMs, the capacity to detect and repair fresh discrepancies depends on a distributed
corrective architecture. Grounding may establish what an output is about; that architecture
determines whether a fresh discrepancy can prompt revision.

\ifttpanonymous\else
\section*{Acknowledgements}
I thank Geoffrey K. Pullum for comments on coherence theory and the history of
\mention{true}.
\fi

\section*{Supplementary information}
\textbf{Online Resource 1: Proposed empirical designs.} Detailed specifications for the
crossed clinical design, checker-independence comparison, and temporal test of live
answerability.

\clearpage
\printbibliography

\end{document}